\documentclass[wcp]{jmlr}

\usepackage{longtable}

\usepackage{booktabs}
\usepackage{amssymb}
\usepackage{amsmath}
\usepackage{graphicx}
\usepackage{bbm}
\usepackage{enumitem}
\usepackage{cases}
\usepackage{url}
\usepackage{sistyle}
\usepackage{xcolor}
\usepackage{bm}

\jmlrvolume{80}
\jmlryear{2017}
\jmlrworkshop{ACML 2017}

\title[Locally Smoothed Neural Networks]{Locally Smoothed Neural Networks}

\author{\Name{Liang Pang${}^{\dag\ast}$} \Email{pl8787@gmail.com}\\
  \Name{Yanyan Lan${}^{\dag}$} \Email{lanyanyan@ict.ac.cn}\\
  \Name{Jun Xu${}^{\dag}$} \Email{junxu@ict.ac.cn}\\
  \Name{Jiafeng Guo${}^{\dag}$} \Email{guojiafeng@ict.ac.cn}\\
  \Name{Xueqi Cheng${}^{\dag}$} \Email{cxq@ict.ac.cn}\\
  \addr $\dag$CAS Key Lab of Network Data Science and Technology, Institute of Computing Technology, \\ Chinese Academy of Sciences, Beijing, China\\
  \addr $\ast$University of Chinese Academy of Sciences, Beijing, China}

\editors{Yung-Kyun Noh and Min-Ling Zhang}

\begin{document}

\maketitle

\begin{abstract}
Convolutional Neural Networks (CNN) and the locally connected layer are limited in capturing the importance and relations of different local receptive fields, which are often crucial for tasks such as face verification, visual question answering, and word sequence prediction. To tackle the issue, we propose a novel locally smoothed neural network (LSNN) in this paper. The main idea is to represent the weight matrix of the locally connected layer as the product of the kernel and the smoother, where the kernel is shared over different local receptive fields, and the smoother is for determining the importance and relations of different local receptive fields. Specifically, a multi-variate Gaussian function is utilized to generate the smoother, for modeling the location relations among different local receptive fields. Furthermore, the content information can also be leveraged by setting the mean and precision of the Gaussian function according to the content. Experiments on some variant of MNIST clearly show our advantages over CNN and locally connected layer.
\end{abstract}

\section{Introduction}
\label{introduction}
Recently, Convolutional Neural Networks (CNN)~\cite{lecun1998gradient,krizhevsky2012imagenet} have attracted great attentions in the communities of computer vision, speech recognition, and natural language processing. One advantage of CNN is that it has the ability to exploit the local translational invariance, through adopting the local connectivity and weight sharing strategies. For example, in the traditional image recognition problem, since the precise location of a feature is independent of the class labels, the weight sharing strategy can benefit from the location invariant features~\cite{lecun1989generalization}.

When facing more challenging tasks such as Face Verification~\cite{taigman2014deepface}, Visual Question Answering (VQA)~\cite{xu2015ask,chen2015abc}, and Word Sequence Prediction~\cite{wang2015gen}, weight sharing has shown its limitations in distinguishing the importance of different local receptive fields. Taking VQA for example, given a text question ``\emph{What's on the table?}'', it is better if the model could automatically focus on the local receptive fields that related to the objects above the ``\emph{table}'' and ignore others, i.e.~the weights of corresponding local receptive fields should be larger than others.

The locally connected layer \cite{gregor2010emergence,taigman2014deepface,huang2012learning,wang2015gen}, which assigns a free weight to each local receptive field, can partially address the problem. However, the generalization ability will be hurt, since it fails to model the relationships among multiple local receptive fields: (1) from the viewpoint of the location, the local receptive fields with respect to overlapping/nearby pixels should have some relationships, since the images are by natural locally smooth and the neighbouring pixels have more correlation than others~\cite{bruna2013spectral}; (2) from the viewpoint of the content, the objects are usually spatially continuous, therefore, local receptive fields with similar pixels would have similar importance. Follow the unrolling strategy~\cite{chellapilla2006high}, in locally connected layer, corresponding kernel weights for each local receptive field are concatenated into a matrix, namely the weight matrix. 

\begin{figure}
  \centering
  \includegraphics[width=9.0cm]{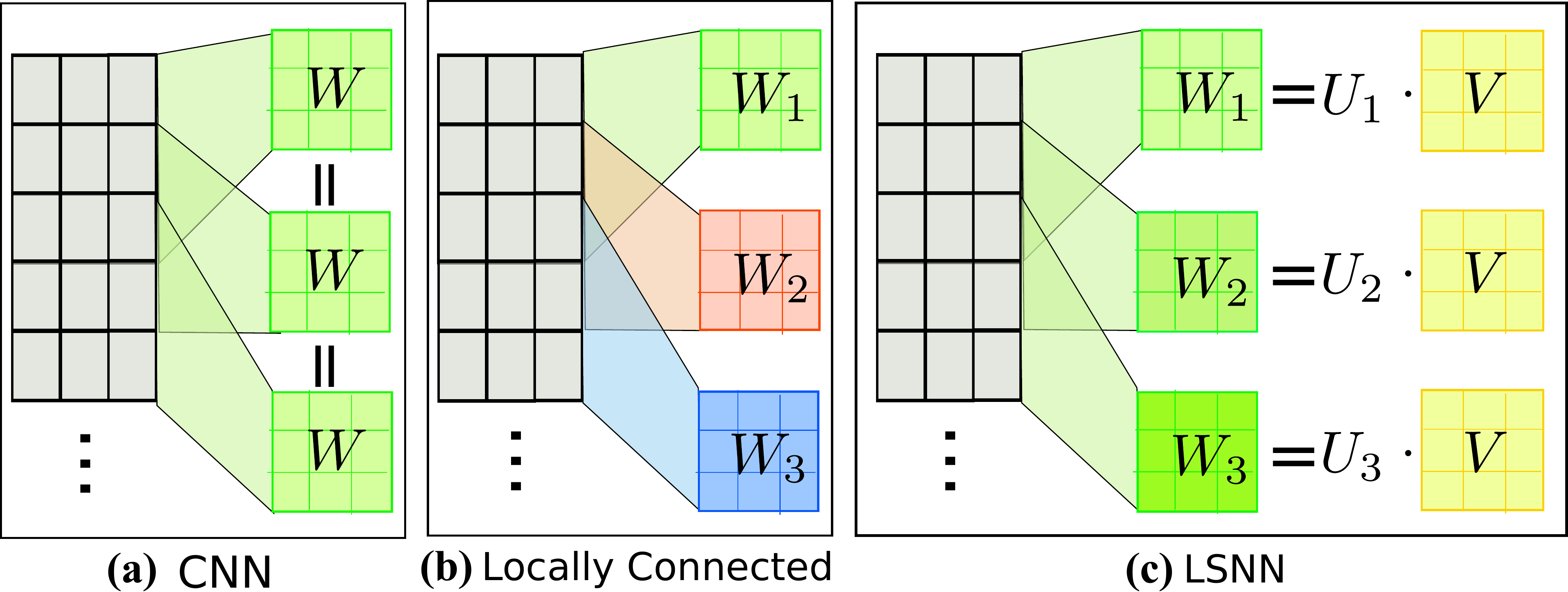}\\
  \caption{Relations of CNN, locally connected and LSNN.}\label{fig:kernel}
\end{figure}

To address the above issues, we propose a novel locally smoothed neural network (LSNN) in this paper. Different from the weight sharing strategy in CNN (as shown in Figure~\ref{fig:kernel}(a)) and the weight free strategy in locally connected layer (as shown in Figure~\ref{fig:kernel}(b)), we propose to use a weight smoothing strategy in LSNN (as shown in Figure~\ref{fig:kernel}(c)). 
Specifically, the weight matrix defined in locally connected layer, is represented as the product of the smoother and the kernel, where each row vector of the weight matrix stands for the weight parameters $W_i$ of a local receptive fields. The kernel is a vector similar to that in CNN. The smoother is a vector in which each element indicates the importance of the local receptive field at the corresponding position.

In order to capture the location relations among different local receptive fields, i.e.~neighbouring fields should be similar, a multi-variate Gaussian function is used to generate the smoother. Specifically, a multi-variate Gaussian function generates a multi-variate density object (e.g., a matrix in the case of two dimensions). Then the smoother vector is obtained by setting each value in the multi-variate density object to the position of the corresponding local receptive field in the vector.
Furthermore, the content information can be utilized to determine the importance of different local receptive fields, by setting the mean and precision of the Gaussian function according to the content. For example, a neural network with fully connected regression layer can be utilized to generate the parameters of Gaussian from the input data.

Our analysis shows that LSNN has a close relation with CNN and the locally connected layer, as illustrated in Figure~\ref{fig:kernel}. Specifically, CNN can be viewed as an LSNN with identical smoother. On the other hand, the locally connected layer can be approximated by an LSNN with free smoother. Furthermore, LSNN is equivalent to the locally connected layer if the rank of the weight matrix equals to 1. Therefore, LSNN makes a balance between CNN and the locally connected layer. The experiments on a series of designed MNIST show that LSNN outperform CNN and the locally connected layer.

Main contributions of the paper include that:

1) the proposal of a novel locally smoothed neural network, in which a smoother is introduced to capture the importance and relations of different local receptive fields;

2) the introduction of a multi-variate Gaussian function as the smoother to leverage the `where' and `what' information in determining the importance and relations of different local receptive fields;

3) the experimental analysis on a series of designed MNIST demonstrates our advantages over CNN and the locally connected layer.

\section{Backgrounds}
\label{backgrounds}
We give an introduction to the convolutional neural networks and locally connected layer.
\subsection{Convolutional Neural Networks}
A convolutional neural network is a type of feed-forward artificial neural network where the individual neurons are tiled in such a way that they respond to overlapping regions in the visual field, called local receptive fields. CNNs have the following two distinguishing feature:

(1) Local connectivity: following the concept of receptive fields, CNNs exploit spatially local correlation by enforcing a local connectivity pattern between neurons of adjacent layers. The architecture thus ensures that the learnt `filters' produce the strongest response to a spatially local input pattern.

(2) Weight sharing: each filter is replicated across the entire visual field in CNN. These replicated units share the same parameterization (weight vector and bias) and form a feature map. This means that all the neurons in a given convolutional layer detect exactly the same feature. Replicating units in this way allows for features to be detected regardless of their position in the visual field, thus constituting the property of translation invariance.

Mathematically, given the input $X$ (an image for example), locally connectivity requires that each filter is conducted on a local receptive fields, i.e.~a patch vector $X_p$, where $p$ stands for the position of the patch in the whole image.
According to weight sharing, the activation (feature map) of this local receptive fields is produced by compositing the patch with a shared kernel $W$, i.e.~$a_p=X_p^\top W$.

Thanks to the nice property of local connectivity and weight sharing, CNNs have gained significant performance improvements in many fundamental problems such as image recognition and classification. However, they are not perfect for many recently proposed tasks, due to the limitation of weight sharing in capturing the importance of different locations~\cite{xu2015ask,chen2015abc,taigman2014deepface} .
\subsection{Locally Connected Layer}
The locally connected layer~\cite{lecun1989generalization} keeps the property of local connectivity in CNN, while assumes the filters with different local receptive fields to be independent with each other. As shown in \cite{gregor2010emergence}, it is more biologically plausible to remove the assumption of identical replicas of filters.

Formally, the activation (feature map) of this local receptive fields is produced by compositing the patch with an independent kernel $W_p$, i.e.~$a_p=X_p^\top W_p$.

Benefiting from the relaxation of weight sharing in \mbox{CNN}, the locally connected layer has the ability to model some more complicated tasks such as Visual Question Answering (VQA), Face Verification, and Word Sequence Prediction. However, the number of parameters increases dramatically from $k$ to $m\times k$ where $m \gg k$, where $m$ denotes the number of patches. Therefore, a very large-scale labeled dataset (about 4.4 million for the task of face verification) is usually required to train the locally connected layer, as mentioned in \cite{taigman2014deepface}. More importantly, the locally connected layer fails to model the relations of parameters in different locations, which limits the further improvements of the model on aforementioned tasks.
\section{Locally Smoothed Neural Networks}
\label{model}
To tackle these issues, we propose a novel locally smoothed neural network (LSNN for short).
From the above analysis, the main problem of existing methods lies in that the constraint of weight sharing is too strict in CNN, while the assumption of parameter independence in different local receptive fields is too free in the locally connected layer. Therefore, a natural idea is to use a weighted kernel over different local receptive fields.

\begin{figure}
  \centering
  \includegraphics[width=8cm]{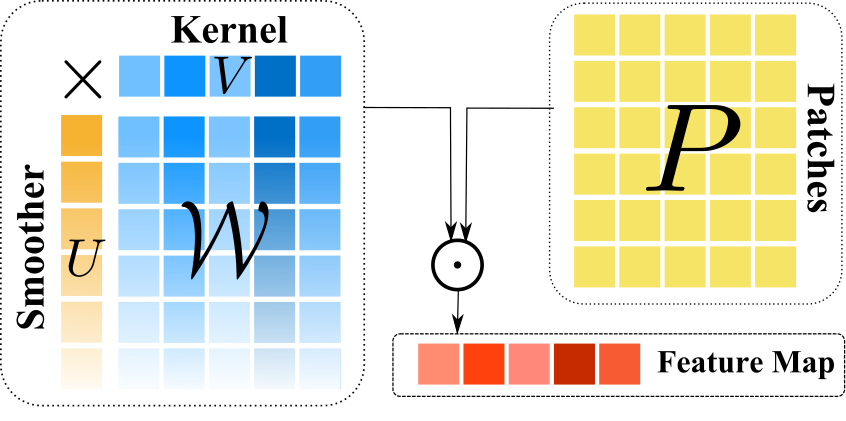}\\
  \caption{The structure of LSNN.}\label{mechanism}
\end{figure}

\subsection{Network Architecture}
The structure of LSNN is illustrated in Figure~\ref{mechanism}.
Specifically, we first put all weight parameters of different local receptive fields together to form a weight matrix, i.e.~$\mathcal{W}=[W_p]_{p\in\mathcal{P}}$, where $W_p$ stands for each kernel vector in different location $p$, and $\mathcal{P}$ stands for the collection of all possible locations. 
This can be achieved by the unrolling strategy as illustrated in \cite{chellapilla2006high}. Taking one dimensional inputs as an example, given the input $X=[x_1, x_2, \dots, x_n]$, the $i$-th patch $P_i$ with dimension $k$ is defines as\footnote{If padding is further considered, the range of $i$ will become $i=1,\cdots,n$.} $P_i=[x_{i}, x_{i+1}, \dots, x_{i+k-1}]^\top, i=1,\cdots,m$ where $m=n-k+1$. The corresponding kernel is denoted as $W_i$. Therefore, the weight matrix can be represented as a vector $\mathcal{W}=[W_1,W_2,\cdots,W_{m}]^\top$.

It is obvious that there exist some relations among the \mbox{kernels} with neighbouring/similar local receptive fields. For example, the patch $P_i$ and patch $P_{i+1}$ have $k-1$ overlapping elements. These overlapped elements may play similar role in the recognition or classification process. In this paper, we define a smoothed kernel to model the relations.
\begin{figure}
  \centering
  \includegraphics[width=8cm]{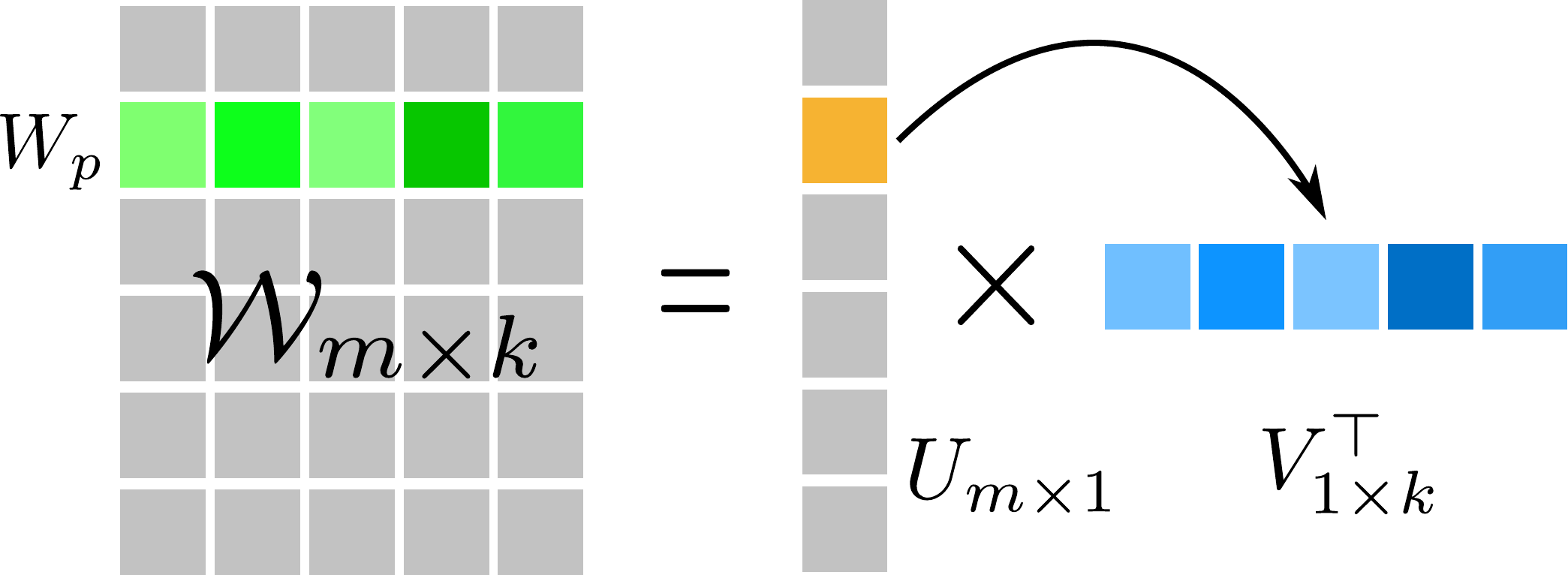}\\
  \caption{Illustration of the moother and the kernel.}\label{fig:factor_1}
\end{figure}

Specifically, we consider the factorization of the $m\times k$ weight matrix, defined as follows.
\begin{equation}
\label{eq:factor}
\mathcal{W}=UV^\top,\,\,\text{i.e.}~W_p=U_pV^\top,\forall p\in\mathcal{P},
\end{equation}
where $U$ is a $m\times 1$ vector with each element stands for the importance of different filter, named {\em smoother}, and $V$ is the $k\times 1$ {\em kernel} vector as illustrated in Figure~\ref{fig:factor_1}.

From Eq.~(\ref{eq:factor}), we can see that LSNN views the weight parameters of each filter as a weighted kernel, with the weight related to the corresponding patch. Therefore, the location and content information can be leveraged to determine the weights for different filters. In this way, LSNN makes a balance between CNN and the locally connected layer.

\subsection{Gaussian Smoother}
\label{apl}
In this paper, we introduce a Gaussian smoother to capture the `where' and `what' information of the data.
Gaussian function is defined with two parameters: the mean $\bm{\mu}$ which determine the position and precision $\bm{\Lambda}$ which controls the area. Therefore the smoother can be generated as the following two steps. The Gaussian parameters $\bm{\mu}$ and $\bm{\Lambda}$ are first generated from the input patches $X_p$, then the smoother $U$ is generated from the parameters $\bm{\mu}$ and $\bm{\Lambda}$:
\begin{equation}
    \begin{aligned}
	  &\bm{\mu} = g(X), \ \ \ \bm{\Lambda} = h(X), \ \ \ U= f(\bm{\mu}, \bm{\Lambda}),
    \end{aligned}
\end{equation}

where function $g$ and $h$ are the feed forward neural neural networks for generating the parameter $\bm{\mu}$ and $\bm{\Lambda}$, called Parameter Network in this paper. Function $f$ is a Gaussian function, and for each position $p$ in $U$, the value $U_p$ is calculated as:
\begin{equation}
    \label{forward}
    U_p =\exp(-(p-\bm{\mu})^\top \bm{\Lambda} (p-\bm{\mu})).
\end{equation}

\begin{figure}
  \centering
  \includegraphics[width=9cm]{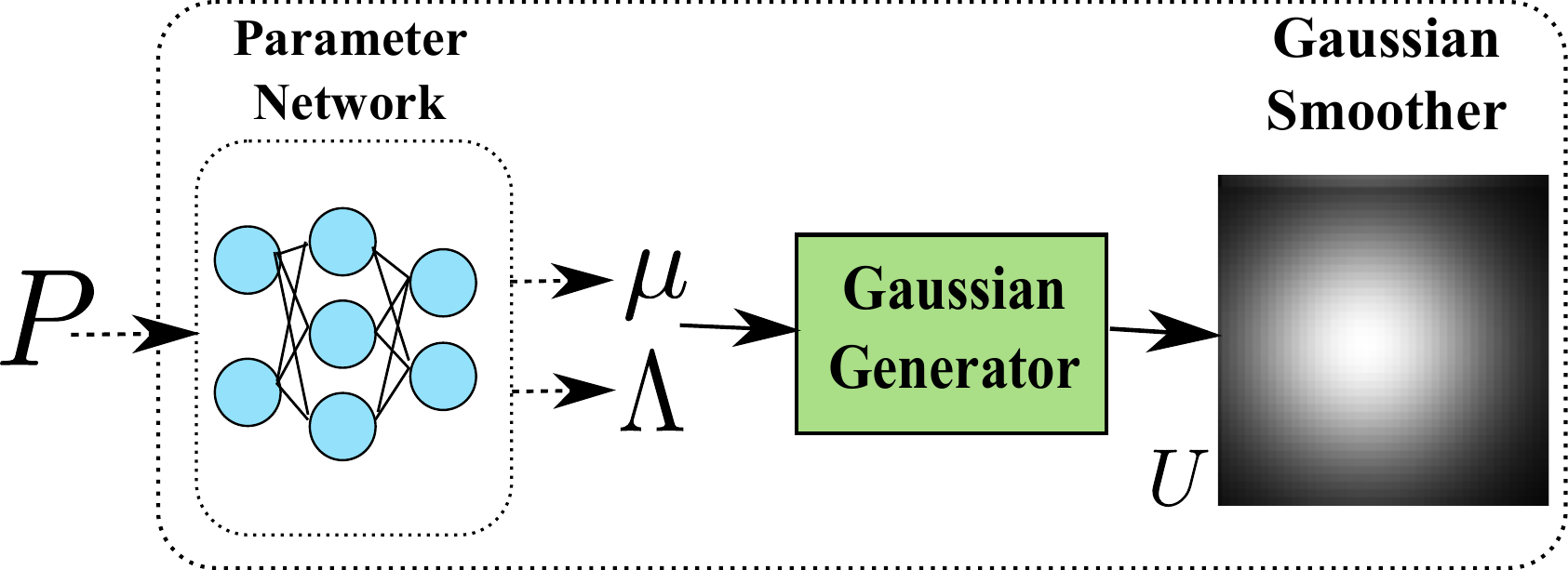}\\
  \caption{An example of two dimensional Gaussian smoother.}
  \label{gaussian_smoother}
\end{figure}

If the function $g$ and $h$ are degenerative, i.e.~$\bm{\mu}$ and $\bm{\Lambda}$ are free parameters learned from data, the smoother can capture the location relations among different filters. Otherwise the content information can be further utilized in determine the importance of different filters. For example, we use a two layer convolutional neural network with a fully connected layer for regression as the Parameter Network in our experiments as illustrated in Figure~\ref{gaussian_smoother}.

In this paper, we give a two dimensional Gaussian smoother as an example to show the generation of $U$.

The following notations are first listed to represent the input and parameters of the Gaussian function.
\begin{equation}
    p = [p_1, p_2]^\top, \ \ \ \bm{\mu} = [\mu_1, \mu_2]^\top, \ \ \ \bm{\Lambda} =
    \left [
    \begin{aligned}
        &\lambda_1 \ \ \lambda_2 \\
        &\lambda_2 \ \ \lambda_3
    \end{aligned}
    \right ],
\end{equation}
where $p_1, p_2, \mu_1, \mu_2 \in \mathbb{R}$ and $\bm{\Lambda}$ is constrained to be a symmetric positive defined matrix.

\subsection{Optimization}
The optimization is relatively straightforward with the standard back-propagation (BP)~\cite{williams1986learning} algorithm.
Here we take two dimensional Gaussian smoother as an example to demonstrate the BP process.

Since it is usually hard to keep the constraint of $\bm{\Lambda}$ during the net optimization process by using back propagation algorithm, we turn to optimize a free parameter $\bm{\Phi}$ instead. Because any positive defined matrix $\bm{\Lambda}$ can be interpreted as the product of two symmetric matrix $\bm{\Phi}$, shown as \mbox{Eq.~(\ref{lambda_fact})}.
\begin{equation}
    \label{lambda_fact}
    \begin{aligned}
        \bm{\Lambda}\!=\! \bm{\Phi}^\top \bm{\Phi} \!=\!
        \left [
        \begin{aligned}
            &\alpha \ \ \gamma \\
            &\gamma \ \ \beta
        \end{aligned}
        \right ]\!
        \left [
        \begin{aligned}
            &\alpha \ \ \gamma \\
            &\gamma \ \ \beta
        \end{aligned}
        \right ]
        \!=\!
        \left [
        \begin{aligned}
            &\alpha^2\!+\!\gamma^2 \ \ (\alpha\!+\!\beta)\gamma \\
            &(\alpha\!+\!\beta)\gamma \ \ \beta^2\!+\!\gamma^2
        \end{aligned}
        \right ].
    \end{aligned}
\end{equation}
where $\alpha$, $\beta$ and $\gamma\in \mathbb{R}$ are parameters to determine an \mbox{unique} matrix $\bm{\Lambda}$.

Based on the above notations, we give the BP process for calculating the gradient of parameters $\bm{\mu}$ and $\bm{\Phi}$, which are further used to update the weights in the parameter net\footnote{We did not include the updates of these weights in this paper, since they are typical BP processes as used in MLP/CNN.}.

The original forward process Eq.~(\ref{forward}) can be rewritten as the following equation:
\begin{equation}
    \label{forward2d}
    U_{p} = \exp(-(p-\bm{\mu})^\top \bm{\Phi}^\top \bm{\Phi} (p-\bm{\mu})).
\end{equation}

Then the backward process of calculating the gradient can be written as:
\begin{equation}
    \label{backward2d}
    \begin{aligned}
        \partial U_p / \partial \bm{\mu} &= 2 U_p \bm{\Phi}^\top \bm{\Phi} (p-\bm{\mu}) , \\
        \partial U_p / \partial \bm{\Phi} &= -2 U_p (p-\bm{\mu}) (p-\bm{\mu})^\top \bm{\Phi}.
    \end{aligned}
\end{equation}

Please note that we normalize each dimension of position to $[0,1]$ in LSNN, similar to the coordinate normalization strategy as used in~\cite{jaderberg2015spatial}, . Taking two dimension as an example, $p=[p_1,p_2]$ will be normalized to $0\leq p_1\leq 1$ and $0\leq p_2\leq 1$. The advantage of conducting such normalization lies in that the optimization difficulty with large parameters will be avoided and the convergence rate can be improved. However, this kind of Gaussian function may meet an ill-posed problem that the parameter $\bm{\mu}$ located far from the image, e.g.~$\bm{\mu} \ll 0$ or $\bm{\mu} \gg 1$. In this case, all input will be screened, and no valid gradient will be back propagated. To address this problem, we further propose a normalized Gaussian smoother, in which the parameters of mean will be constrained to the window of the input image data, described as follows. In this way, the gradient will be amplified when $\bm{\mu}$ moves out of the $\mathcal{P}$.

In the forward step $U_p$ is normalized as follows.
\begin{equation}
    \widehat{U_p} =  U_p / \sum_{p' \in \mathcal{P}} U_{p'} .
\end{equation}

In the backward step, Eq.(\ref{backward2d}) will become the following \mbox{form}:
\begin{equation}
    \begin{aligned}
        \frac{\partial \widehat{U_p}}{\partial \bm{\mu}} &= \frac{ \frac{\partial U_p}{\partial \bm{\mu}} \sum_{p' \in \mathcal{P}} U_{p'} - U_p \sum_{p' \in \mathcal{P}} \frac{\partial U_{p'}}{\partial \bm{\mu}} }{ (\sum_{p' \in \mathcal{P}} U_{p'})^2 } , \\
        \frac{\partial \widehat{U_p}}{\partial \bm{\Phi}} &= \frac{ \frac{\partial U_p}{\partial \bm{\Phi}} \sum_{p' \in \mathcal{P}} U_{p'} - U_p \sum_{p' \in \mathcal{P}} \frac{\partial U_{p'}}{\partial \bm{\Phi}} }{ (\sum_{p' \in \mathcal{P}} U_{p'})^2 }
    \end{aligned}.
\end{equation}
For implementation, we use the Stochastic Gradient Decent (SGD) with momentum method to train our models. The learning rate of the Parameter Network is set to one tenth of the other parts of LSNN, since the parameters of the Gaussian smoother are only a small part of the whole parameters in LSNN. If  the learning rate is large, the back propagation errors will accumulate on these parameters during the training process, leading to the problem of slow convergence.

\section{Discussion}
\label{theory}
In this section, discuss the relations of LSNN with previous models and its relations with the attention mechanism.
\subsection{Relations with CNN and Locally Connected Layer}
LSNN has a close relationship with CNN and locally connected layer.

a. \emph{CNN is a special case of LSNN by setting the smoothers to a vector of ones.}

Based on the above introduction, LSNN views the weight matrix as the product of two vectors, i.e.~the smoother and the kernel, described as the following equation.
\begin{equation}\label{eq:theoretical}
	\mathcal{W}=UV^\top,
\end{equation}
where $\mathcal{W}$ is the $m\times k$ weight matrix, $U$ is a $m\times 1$ vector, and $V$ is a $k\times 1$ vector.

If $U=\mathcal{I}$, a vector of ones, i.e.~$\mathcal{I}=[1,1,\cdots,1]^\top$, the above equation will become the following form:
\begin{equation}
\mathcal{W}=\mathcal{I}V^\top,\,\,\text{i.e.}~W_p=V^\top,
\end{equation}
Therefore, $W_p$ is the same for different positions, and a weight sharing effect is achieved. Therefore, LSNN degenerates to a CNN.

b. \emph{LSNN can be viewed as an approximation of locally connected layer. Furthermore, if the rank of the weight matrix is 1 in the locally connected layer, LSNN is equivalent to the locally connected layer.}

As described above, the locally connected layer removes the constraint of identical filters in CNN and resorts to set the weight parameters independently in different local receptive fields. Therefore, the weight matrix $\mathcal{W}$ is totally free, with each kernel represented as $W_p$.

Since LSNN can be viewed as factorizing the weight matrix of the locally connected layer into the product of two vectors, LSNN approximates the weight matrix of the locally connected layer $W_p$ with $U_pV^\top$ at each position.
The approximation accuracy heavily depends on the rank of the weight matrix in the locally connected layer.
According to the property that every finite-dimensional matrix has a rank decomposition, LSNN is equivalent to the locally connected layer when the rank of the weight matrix is equal to 1.

From the analysis, we can see that these three neural networks are making different assumptions on the filters over different local receptive fields:
CNN assumes the identical filter, which is the most strict constraint; the locally connected layer assumes a free filter, which is the most flexible;
while LSNN is makes a balance between them, which assumes that all the filters can be represented as a weighted kernel, and these filters are only using different weights determined by the corresponding location or content. Intuitively, the soft smoother works as a `regularization', and the generalization performance will be improved in this way. 

\subsection{Extensions of LSNN}
LSNN is a general framework, and the main idea lies in that the weight matrix is represented as the product of the smoother and the kernel. Therefore, several extensions can be made within this flexible framework. Specifically, we introduce the extensions beyond the Gaussian smoother and higher dimensional factorization.
\subsubsection{Beyond Gaussian Smoother}
Gaussian function is chosen as the smoother in this paper, because it is not only unimodel and continuous function that easy for analyse and optimization, but also appropriate for modeling the location relations of different filters. In practice, we can also choose other smoothers.

Facing with more complicated applications, some more complicated functions can be utilized as the smoother to adapt the requirements.
For example, in many \mbox{applications} such as scene detection and sentiment analysis, two or more disconnected parts are usually jointed in determining the label. For example, there are usually multiple objects in scene detection, such as pedestrian, cars, and flags, et al. In sentiment analysis, the final sentiment is usually determined by the composition of several emotional words. In this case, the usage of Gaussian function will concentrate on the centre of the object, thus hurt the final performance. A more reasonable way is to replace the Gaussian function with Mixture Gaussian function. Consequently, the model can produce multiple centres and solve the above problem.

\subsubsection{Beyond One Dimensional Factorization}
\begin{figure}
  \centering
  \includegraphics[width=8cm]{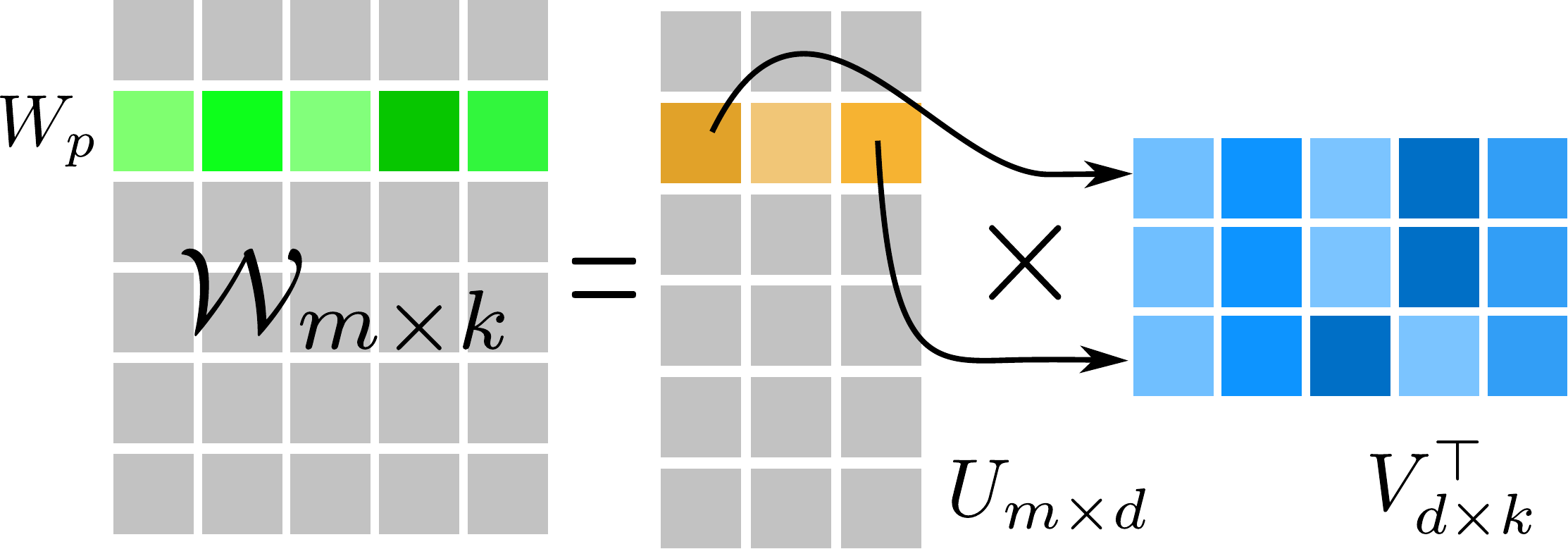}\\
  \caption{High dimensional factorization.}\label{fig:factor_d}
\end{figure}

In LSNN, the $m \times k$ weight matrix $\mathcal{W}$ is represented as the product of two vectors ($UV^\top$) in this paper. That is to say, we are conducting a one dimensional factorization. Actually, high dimensional factorization can also be utilized.
As shown in Figure~\ref{fig:factor_d}, assume that $U$ is a $m\times d$ matrix and $V$ is a $k\times d$ matrix, $d\geq 2$. Then the two matrices can be represented as follows.
\begin{equation*}
U=[U^{(1)},\cdots,U^{(m)}]^\top,\,\,V=[V^{(1)},\cdots,V^{(d)}],
\end{equation*}
where $V^{(i)}$ (with dimension $k \times 1$) stands for the $i$-th vector in $V$ and represents the $i$-th kernel. $U^{(j)}$ (with dimension $d \times 1$) stands for the $j$-th vector in $U$ and represents the corresponding weight for $d$ kernels at position $j$.
In this case, the kernel $W_p$ turns to be the linear combination of $d$ kernels in $V$, described as follows.

\begin{displaymath}
W_p=\sum_{l=1}^d U_l^{(p)}V^{(l)},
\end{displaymath}
where $W_p$ is settled in the $p$-th row of the weight matrix $\mathcal{W}$, and $U_l^{(p)}$ stands for the $l$-th value of the vector $U^{(p)}$.
We can see that when $d=1$, filters at different locations are linear correlation. While when $d>1$, the filters are different linear combinations of multiple kernels. This is more flexible to model the relations among different local receptive fields.

\subsection{Relations with the Attention Mechanism}
Recently, the attention mechanism has attracted much attention in the area of deep learning~\cite{cho2015describing,gregor2015draw,larochelle2010learning,ba2014multiple}. In general, attention mechanisms are components of prediction systems that allow the system to sequentially focus on different subsets of the input. The selection of the subset is typically conditioned on the state of the system, which is itself a function of the previously attended \mbox{subsets}. Our LSNN can be viewed as a \emph{soft attention} mechanism, which avoids a hard selection of subsets to attend, and uses a soft weighting for the different subsets instead. Furthermore, the Gaussian smoother takes both the location and content information into consideration, which allows the system to focus on the distinct aspects of the input. Therefore, the ability of extracting the most relevant information for each piece of the output can be enhanced, yielding improvements in the quality of the generated outputs.

\section{Experiments}
\label{experiments}
We conduct experiments~\footnote{The source code of LSNN attaches in \url{https://gitlab.com/pl8787/CaffeAttention}.} on some variant of MNIST to evaluate the performances of LSNN.
Specifically, the first experiment on cluttered translated MNIST shows that LSNN has the ability to denoise and to focus on the desired concentration.
In Section~\ref{exp_part2}, we conduct experiments on intention signed MNIST, in order to show that LSNN can distinguish the important local receptive fields by recognizing the signs and the implicit intention automatically with a content based smoother.
Section~\ref{exp_part3} demonstrates the experimental results on cluttered MNIST sequence, showing that LSNN has the ability of capturing more complicated spacial distribution by using multiple smoothers.

CNN and the locally connected layer are chosen as the baselines. For CNN, two convolutional layers and max-pooling layers are used, with 16 kernels and 32 kernels respectively. Then one fully connected layer with 256 hidden nodes and a dropout layer~\cite{hinton2012improving} with 0.5 ratio are attached for the classification.
LSNN and the locally connected layer have similar structures with CNN, with differences lie in the first convolutional layer. For Locally Connected Layer, the first convolutional layer is replaced by locally connected layer. For LSNN, 16 Gaussian smoothers are used for 16 kernels in the first layer. There are two versions of LSNN, according to different Gaussian smoother. When the parameters in Gaussian are free, we denote it as LSNN-Location. When the parameters are generated by a parameter net, we denote it as LSNN-Content.

\subsection{Experiments on Cluttered Translated MNIST}
\label{exp_part1}

In this experiment, we evaluate the effectiveness of LSNN in denoising and concentration. The data is based on MNIST and generated as follows (similar to \cite{mnih2014recurrent}): firstly, each image, a $28 \times 28$ pixels MNIST digit, is placed in a uniform random location in an $84 \times 84$ black background image. We then add 32 random $6 \times 6$ crops sampled from the original MNIST training dataset, creating distractor. Finally the image is resized from $84 \times 84$ to $42 \times 42$. We can see that the noise will occupy 89\% of an image averagely.

\begin{table}[!htbp]
  \centering
  \caption{The error rate on the clutter translated MNIST task.}\label{Table:Clutter}
  \begin{tabular}{l l}
     \hline
     Model & Error \\
     \hline
     Convolutional Layer & 3.88\% \\
     Locally Connected Layer & 9.54\% \\
     LSNN-Location & 3.08\% \\
     LSNN-Content & \textbf{2.89\%} \\
     \hline
   \end{tabular}
\end{table}

We apply CNN, the locally connected layer and LSNN on this data. The results are shown in Table~\ref{Table:Clutter}. We can see that LSNN outperforms CNN and the locally connected layer.
Among the two versions of LSNN, LSNN-Content performs slightly better than LSNN-Location.
Therefore we can conclude that leveraging the content information in defining the smoother is helpful for denoising and thus can improve the performance.

The results indicate that the locally connected layer is sensitive to the noise, since the parameters are totally free.
However, the constraint of weight sharing in CNN works as a good `regularization', yielding a better generalization than the locally connected layer.

\begin{figure}
  \centering
  \subfigure[] { \label{fig:example_origin}
    \includegraphics[width=0.10\columnwidth]{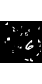}
  }
  \subfigure[] { \label{fig:example_gaussian}
    \includegraphics[width=0.20\columnwidth]{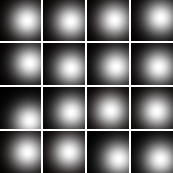}
  }
  \subfigure[] { \label{fig:example_attention}
    \includegraphics[width=0.10\columnwidth]{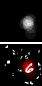}
  }
  \caption{(a) Image of a number ``\emph{6}'' with additional cluttered noise. (b) The 16 Gaussian smoothers. (c) Top: blending of these 16 Gaussian smoothers. Bottom: The image with important area identified by LSNN, shown in red.}\label{example}
\end{figure}

We further conduct an experiment to visualize the ability of LSNN in distinguishing the important local receptive fields from others. We choose LSNN-Content as an example to conduct the experiments, and the results are shown in Figure~\ref{example}. Figure~\ref{fig:example_origin} shows an image of number ``\emph{6}'' with additional cluttered noise. The learned 16 Gaussian smoothers by LSNN are shown in Figure~\ref{fig:example_gaussian}, where the brightness stands for the strength. The blending of these 16 Gaussian smoothers are shown in the top of Figure~\ref{fig:example_attention}, which is used to identify the concentrating area. We can see that the important area identified by LSNN (i.e.~red area) perfectly covers the area that the number `\emph{6}' lies in. In other words, the important area for recognizing number `\emph{6}' is automatically identified and other areas (noise) are eliminated.

\begin{figure}
  \centering
  \includegraphics[width=7cm]{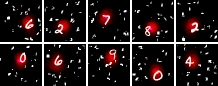}
  \caption{Visualization of LSNN on more example images.}\label{fig:sch4}
\end{figure}

More examples are shown in Figure~\ref{fig:sch4}. Please note that the results of LSNN-Location are similar, the only difference is that the Gaussian smoothers are identical across all images, since the parameters of Gaussian are independent of the content. We omits the results due to the space limitation.

\subsection{Experiments on Intention Signed MNIST}
\label{exp_part2}
We also conduct experiment to demonstrate the influence of the content on LSNN, for distinguishing important local receptive fields.
The data is generated as follows: firstly, two $28 \times 28$ pixels MNIST digits are placed in a $84 \times 84$ black background image. Then add a sign to identify the number we want to classify. Two kinds of signs are used in this paper. The first one is an arrow to indicate the digit we need to recognize, namely Arrow Intention (denoted as Arrow Int. in Table~\ref{Table:Signed1}). The other one is a rectangle, which is located around the digit we need to recognize, namely Rectangle Intention (denoted as Rect Int. in Table~\ref{Table:Signed1}). The two kinds of signs introduce different spatial distributions. In the scenario of Arrow Intention, digits are only placed in the four corners of the image and four types of arrow are used to point out the target digit. Furthermore, in the case of Rectangle Intention, digits are placed randomly, and the target digit locates in the rectangle. 
Finally, the images are also resized from $84 \times 84$ to $42 \times 42$, the same as in the first experiment. The goal is to predict the digit that the signs pointed to.

\begin{table}[!htbp]
  \centering
  \caption{The error rate on the intention signed MNIST task.}
  \label{Table:Signed1}
  \begin{tabular}{l l l}
     \hline
     Model & Arrow Int. & Rect Int. \\
     \hline
     Convolutional Layer & 2.05\% & 6.25\% \\
     Locally Connected Layer & 2.53\% & 13.76\% \\
     LSNN-Location & 1.60\% & 4.02\% \\
     LSNN-Content & \textbf{1.25\%} & \textbf{3.33\%} \\
     \hline
   \end{tabular}
\end{table}
\begin{figure}[!htbp]
  \centering
  \includegraphics[width=8cm]{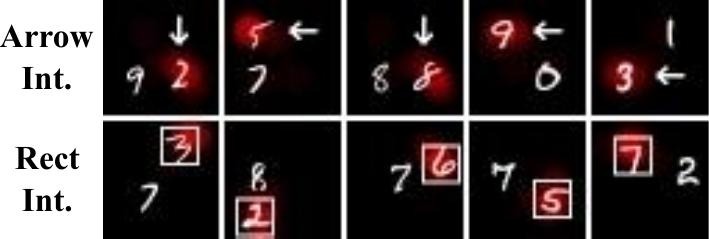}
  \caption{Visualizaton of LSNN on the intent signed MNIST.}\label{fixrnd4sch2direct}
\end{figure}

Table~\ref{Table:Signed1}  shows the experimental results. We can see that LSNN-Location and LSNN-Content outperform the baseline models, and LSNN-Content performs the best. We note that the gap between LSNN-Location and LSNN-Content is larger comparing with cluttered translated experiments in Section~\ref{exp_part1}. That is because the recognition of the correct number requires a content based smoother to distinguish the important local receptive fields that the correct number located.
Our visualization results in Figure~\ref{fixrnd4sch2direct} also verify that LSNN has the ability to distinguish the correct location by the content based Gaussian smoother. Another interesting finding is that the smoother only concentrates on the digital in the arrow intent experiment, regardless of the arrow locations. It indicates that LSNN and the classification CNN work independently in the structure. The LSNN first picks out the desired region by recognizing the arrow, and then CNN recognize the number within the region.

These tasks can also be interpreted as the VQA problem with question `What's the digit that the arrow points to?'/`What's the digit that sets in the rectangle?'.
Therefore, LSNN has the potential of applying to the VQA task, since it has the ability to attend to the area indicated by the question, as shown from the above experimental results.

\subsection{Experiments on Cluttered MNIST Sequence}
\label{exp_part3}
A more interesting question is that whether LSNN has the ability to identify multiple important areas simultaneously, i.e.~there are multiple centers in the image. Moreover, these areas may have some relations.

The cluttered MNIST sequence data is constructed as follows (a little different from \cite{sonderby2015recurrent}): three MNIST digits are placed in a $100 \times 100$ background image. The first digit is chosen randomly, while the following two digits are placed in a right-hand slope of $[-45^\circ, 45^\circ]$. Then the 8 cluttered noise which randomly crops ($6 \times 6$) from the original MNIST training dataset, are added to each image. In the end, the image is resized from $100 \times 100$ to $42 \times 42$. The goal is to predict the 3 digits in the image. The task, thus, becomes a multi-class multi-label classification problem. 

This task is more complicated than the above two tasks shown in Section~\ref{exp_part1} and Section~\ref{exp_part2}. It requires to concentrate on three related locations rather than one. Therefore, we also make a small change to our LSNN to facilitate the study. Specifically, we set 3 independent\footnote{By considering a sequential model such as RNN to capture the relations, the results will be better.} groups of Gaussian smoothers for each filter in the first convolutional layer. Then the output of each group of smoothers will be sent to the same classification network for recognising three digits. Each digit needs to back propagate its own error independently. The parameters of classification network are doubled to tackle the complicated pattern recognition task. The baseline models have the similar structures with LSNN.

\begin{table}
  \centering
  \caption{The error rate on the cluttered MNIST sequence task.}\label{Table:Sequence}
  \begin{tabular}{l l}
     \hline
     Model & Error \\
     \hline
     Convolutional Layer & 23.97\% \\
     Locally Connected Layer & 13.5\% \\
     LSNN-Location & 9.98\% \\
     LSNN-Content & \textbf{6.18\%} \\
     \hline
   \end{tabular}

\end{table}
\begin{figure}
  \centering
  \includegraphics[width=7cm]{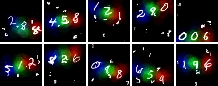}\\
  \caption{Visualization of LSNN on the cluttered MNIST sequence.}\label{seq3}
\end{figure}

The results Table~\ref{Table:Sequence} show that LSNN performs the best in this task. Convolutional Layer performs the worst, simply because convolutional layer can not distinguish the digits in different positions, thus unable to provide divided digit to classification network. Examples shown in Figure~\ref{seq3} illustrate that LSNN can precisely identify the three related important areas automatically (shown as the blue, green, and red, respectively). This is the reason why LSNN can outperform the baseline methods.

\section{Conclusion}
\label{conclusion}
In this paper, we propose a novel LSNN to address the limitations of CNN and the locally connected layer in capturing the importance and relations of different local receptive fields. Our main idea is to factorize the weight matrix of the locally connected layer into the product of two matrix, i.e.~the smoother and the kernel. Analysis shows the relationship of LSNN with previous models: CNN is a special case of LSNN, and LSNN is an approximation of the locally connected layer.  Experimental results on some variations of MNIST data show that: (1) LSNN has the ability to focus on the desired concentration; (2) When the desired concentration is implicitly indicated by some signs in the image, LSNN can also achieve the concentration effect by recognizing the signs and their implicit intention automatically, with a content based smoother; (3) When the concentration has a complicated spatial distribution, e.g.~there are multiple related concentrations, LSNN can effectively capture such distribution by using multiple smoothers.

\acks{This work was funded by the 973 Program of China under Grant No. 2014CB340401, the National Natural Science Foundation of China (NSFC) under Grants No. 61232010, 61433014, 61425016, 61472401, and 61203298, and the Youth Innovation Promotion Association CAS under Grants No. 20144310 and 2016102.}

\bibliographystyle{plain}
\bibliography{attention_cnn}

\begin{thebibliography}{20}
\providecommand{\natexlab}[1]{#1}
\providecommand{\url}[1]{\texttt{#1}}
\expandafter\ifx\csname urlstyle\endcsname\relax
  \providecommand{\doi}[1]{doi: #1}\else
  \providecommand{\doi}{doi: \begingroup \urlstyle{rm}\Url}\fi

\bibitem[Ba et~al.(2014)Ba, Mnih, and Kavukcuoglu]{ba2014multiple}
Jimmy Ba, Volodymyr Mnih, and Koray Kavukcuoglu.
\newblock Multiple object recognition with visual attention.
\newblock \emph{arXiv preprint arXiv:1412.7755}, 2014.

\bibitem[Bruna et~al.(2013)Bruna, Zaremba, Szlam, and LeCun]{bruna2013spectral}
Joan Bruna, Wojciech Zaremba, Arthur Szlam, and Yann LeCun.
\newblock Spectral networks and locally connected networks on graphs.
\newblock \emph{arXiv preprint arXiv:1312.6203}, 2013.

\bibitem[Chellapilla et~al.(2006)Chellapilla, Puri, and
  Simard]{chellapilla2006high}
Kumar Chellapilla, Sidd Puri, and Patrice Simard.
\newblock High performance convolutional neural networks for document
  processing.
\newblock In \emph{Tenth International Workshop on Frontiers in Handwriting
  Recognition}. Suvisoft, 2006.

\bibitem[Chen et~al.(2015)Chen, Wang, Chen, Gao, Xu, and Nevatia]{chen2015abc}
Kan Chen, Jiang Wang, Liang-Chieh Chen, Haoyuan Gao, Wei Xu, and Ram Nevatia.
\newblock Abc-cnn: An attention based convolutional neural network for visual
  question answering.
\newblock \emph{arXiv preprint arXiv:1511.05960}, 2015.

\bibitem[Cho et~al.(2015)Cho, Courville, and Bengio]{cho2015describing}
Kyunghyun Cho, Aaron Courville, and Yoshua Bengio.
\newblock Describing multimedia content using attention-based encoder-decoder
  networks.
\newblock \emph{Multimedia, IEEE Transactions on}, 17\penalty0 (11):\penalty0
  1875--1886, 2015.

\bibitem[Gregor and LeCun(2010)]{gregor2010emergence}
Karo Gregor and Yann LeCun.
\newblock Emergence of complex-like cells in a temporal product network with
  local receptive fields.
\newblock \emph{arXiv preprint arXiv:1006.0448}, 2010.

\bibitem[Gregor et~al.(2015)Gregor, Danihelka, Graves, and
  Wierstra]{gregor2015draw}
Karol Gregor, Ivo Danihelka, Alex Graves, and Daan Wierstra.
\newblock Draw: A recurrent neural network for image generation.
\newblock \emph{arXiv preprint arXiv:1502.04623}, 2015.

\bibitem[Hinton et~al.(2012)Hinton, Srivastava, Krizhevsky, Sutskever, and
  Salakhutdinov]{hinton2012improving}
Geoffrey~E Hinton, Nitish Srivastava, Alex Krizhevsky, Ilya Sutskever, and
  Ruslan~R Salakhutdinov.
\newblock Improving neural networks by preventing co-adaptation of feature
  detectors.
\newblock \emph{arXiv preprint arXiv:1207.0580}, 2012.

\bibitem[Huang et~al.(2012)Huang, Lee, and Learned-Miller]{huang2012learning}
Gary~B Huang, Honglak Lee, and Erik Learned-Miller.
\newblock Learning hierarchical representations for face verification with
  convolutional deep belief networks.
\newblock In \emph{Computer Vision and Pattern Recognition (CVPR), 2012 IEEE
  Conference on}, pages 2518--2525. IEEE, 2012.

\bibitem[Jaderberg et~al.(2015)Jaderberg, Simonyan, Zisserman,
  et~al.]{jaderberg2015spatial}
Max Jaderberg, Karen Simonyan, Andrew Zisserman, et~al.
\newblock Spatial transformer networks.
\newblock In \emph{Advances in Neural Information Processing Systems}, pages
  2008--2016, 2015.

\bibitem[Krizhevsky et~al.(2012)Krizhevsky, Sutskever, and
  Hinton]{krizhevsky2012imagenet}
Alex Krizhevsky, Ilya Sutskever, and Geoffrey~E Hinton.
\newblock Imagenet classification with deep convolutional neural networks.
\newblock In \emph{Advances in neural information processing systems}, pages
  1097--1105, 2012.

\bibitem[Larochelle and Hinton(2010)]{larochelle2010learning}
Hugo Larochelle and Geoffrey~E Hinton.
\newblock Learning to combine foveal glimpses with a third-order boltzmann
  machine.
\newblock In \emph{Advances in neural information processing systems}, pages
  1243--1251, 2010.

\bibitem[LeCun et~al.(1998)LeCun, Bottou, Bengio, and
  Haffner]{lecun1998gradient}
Yann LeCun, L{\'e}on Bottou, Yoshua Bengio, and Patrick Haffner.
\newblock Gradient-based learning applied to document recognition.
\newblock \emph{Proceedings of the IEEE}, 86\penalty0 (11):\penalty0
  2278--2324, 1998.

\bibitem[LeCun et~al.(1989)]{lecun1989generalization}
Yann LeCun et~al.
\newblock Generalization and network design strategies.
\newblock \emph{Connections in Perspective. North-Holland, Amsterdam}, pages
  143--55, 1989.

\bibitem[Mnih et~al.(2014)Mnih, Heess, Graves, et~al.]{mnih2014recurrent}
Volodymyr Mnih, Nicolas Heess, Alex Graves, et~al.
\newblock Recurrent models of visual attention.
\newblock In \emph{Advances in Neural Information Processing Systems}, pages
  2204--2212, 2014.

\bibitem[S{\o}nderby et~al.(2015)S{\o}nderby, S{\o}nderby, Maal{\o}e, and
  Winther]{sonderby2015recurrent}
S{\o}ren~Kaae S{\o}nderby, Casper~Kaae S{\o}nderby, Lars Maal{\o}e, and Ole
  Winther.
\newblock Recurrent spatial transformer networks.
\newblock \emph{arXiv preprint arXiv:1509.05329}, 2015.

\bibitem[Taigman et~al.(2014)Taigman, Yang, Ranzato, and
  Wolf]{taigman2014deepface}
Yaniv Taigman, Ming Yang, Marc'Aurelio Ranzato, and Lars Wolf.
\newblock Deepface: Closing the gap to human-level performance in face
  verification.
\newblock In \emph{Computer Vision and Pattern Recognition (CVPR), 2014 IEEE
  Conference on}, pages 1701--1708. IEEE, 2014.

\bibitem[Wang et~al.(2015)Wang, Lu, Li, Jiang, and Liu]{wang2015gen}
Mingxuan Wang, Zhengdong Lu, Hang Li, Wenbin Jiang, and Qun Liu.
\newblock $ gen $ cnn: A convolutional architecture for word sequence
  prediction.
\newblock \emph{arXiv preprint arXiv:1503.05034}, 2015.

\bibitem[Williams and Hinton(1986)]{williams1986learning}
DE~Rumelhart GE Hinton~RJ Williams and GE~Hinton.
\newblock Learning representations by back-propagating errors.
\newblock \emph{Nature}, pages 323--533, 1986.

\bibitem[Xu and Saenko(2015)]{xu2015ask}
Huijuan Xu and Kate Saenko.
\newblock Ask, attend and answer: Exploring question-guided spatial attention
  for visual question answering.
\newblock \emph{arXiv preprint arXiv:1511.05234}, 2015.

\end{thebibliography}

\appendix





\end{document}